\documentclass[aoas,preprint]{imsart}
\usepackage{amsfonts}
\usepackage{float}

\RequirePackage[OT1]{fontenc}
\RequirePackage{amsthm,amsmath}
\RequirePackage[numbers]{natbib}
\RequirePackage[colorlinks,citecolor=blue,urlcolor=blue]{hyperref}

\arxiv{arXiv:0000.0000}

\startlocaldefs
\numberwithin{equation}{section}
\theoremstyle{plain}
\newtheorem{theorem}{Theorem}

\newtheorem{case}[theorem]{Case}

\newtheorem{conclusion}[theorem]{Conclusion}

\newtheorem{remark}[theorem]{Remark}

\endlocaldefs

\begin{document}
\begin{frontmatter}
\title{Adaptive Scaling}
\runtitle{Adaptive Scaling}
%\thankstext{T1}{Footnote to the title with the ``thankstext'' command.}

\begin{aug}
\author{\fnms{Ting Li}\ead[label=e1]{tlial@connect.ust.hk}},
\author{\fnms{Bingyi Jing}\ead[label=e2]{majing@ust.hk}},
\author{\fnms{Ningchen Ying}
\ead[label=e3]{nying@connect.ust.hk}},
\author{\fnms{Xianshi Yu} 
\ead[label=e4]{xyuai@connect.ust.hk}}

\thankstext{t1}{Some comment}
\thankstext{t2}{First supporter of the project}
\thankstext{t3}{Second supporter of the project}
\runauthor{F. Author et al.}

\affiliation{HKUST}

\address{ Address of First Author:\\
Department of Mathematics \\
The Hong Kong University of Science and Technology,\\
Clear Water Bay, Kowloon, Hong Kong. \\
\phantom{E-mail:\ }\printead*{e1}}

\end{aug}

\begin{abstract}
Preprocessing data is an important step before any data analysis. In this paper, we focus on one particular aspect, namely scaling or normalization. We analyze various scaling methods in common use and study their effects on different statistical learning models. We will propose a new two-stage scaling method. First, we use some training data to fit linear regression model and then scale the whole data based on the coefficients of regression. Simulations are conducted to illustrate the advantages of our new scaling method. Some real data analysis will also be given. 
\end{abstract}



\end{frontmatter}

\section{Introduction}\label{intro}
Nowadays, statistical learning models are widely used to do regression, classification as well as data mining. More and more new learning models are proposed to deal with different types of data sets. Feature Scaling is a necessary step for data preprocessing and is widely used in applications.

The motivation of doing feature scaling is varied. First of all, multi-dimensional data in reality often have different units. For example, we have kilograms for weight, metres for height, years for age and dollars for income. It is necessary to do some feature scaling before we combine all these features together. Secondly, there can be different measurements for each feature.  We can use grams as weight unit, as well as kilograms and tons, but no matter which unit we use, the underlying value does not change. Feature scaling should be applied to prevent the difference of measurement from affecting the final result. Thirdly, there are many different types of data such as numerical data (weight, height, age), categorical data (healthy/sick, white/black/asian/hispanic) and ordinal data (Grades A/B/C/D/E/F, heavy/light/no smokers). It is improper for us to just scale them in the same way. Lastly, some statistical learning algorithms will not perform well if we just use raw data with a wide range of values, while proper feature scaling makes optimal algorithm, such as gradient descent, converges much faster.

Although Standardization\cite{Triola M}, or Z-score, is the most popular scaling method, many other scaling methods are proposed for particular application such as Range Scaling, Pareto Scaling\cite{Eriksson L},  Vast Scaling\cite{Keun HC} and  Level Scaling\cite{Berg R A}. More details are included in the Appendix.

  Although the scaling methods we mentioned above have different formulation, all of them try to eliminate the impact of measurement. However, too many choices lead to another problem, which scaling method should we take? Standardization may be the most popular method followed by Range Scaling, and the other scaling methods are still used in special cases. Generally speaking, which method to pick is mainly determined by what kind of data we have and which statistical learning model do we apply. One safe way is to try all the scaling methods on training data and pick the best one to apply on test data, but this will be time consuming especially for big data problems.
 
We close this section by giving the arrangement of this paper. In section \ref{adaptive}, we propose a new feature scaling method, referred to as "Adaptive Scaling". Scaling affection on models is discussed in section \ref{affection}. Simulation and empirical study are carried out in section \ref{simulation} and section \ref{empirical}. Finally, we conclude this paper and point out some future work in section \ref{conclusion}.

 \section{Adaptive Scaling}\label{adaptive}
In previous sections, we listed some widely used scaling methods including standardization and rescaling. Here we propose a new two-stage scaling method which is called "Adaptive Scaling" and two extension version of it.
\subsection{Adaptive Scaling (AS)}
Assume we have raw data $(X,Y)$ and we separate it into training data $({X}_{training},Y_{training})$ and testing data$({X}_{testing},Y_{testing})$. Then we do Adaptive Scaling on both training data and testing data.

Algorithm:\\
 1. Center Data.
 $$X=X-\bar{{X}}_{training}$$\\
 2. Apply OLS on training data to get coefficients.
 $$\beta = (X_{training}^T X_{training})^{-1}X_{training}^T Y_{training}$$\\
 3. Multiple the coefficients to the raw data.
 $${X'}_i=\beta_i\cdot X_i$$\\
 4. Use ${X'}_{training}$ and $Y_{training}$ to learn the model.\\
 5. Use ${X'}_{testing}$ and $Y_{testing} $ to validate the model.
 
\begin{remark}
Since OLS is the simplest model in statistical learning and it is scale-invarian, AS does feature scaling without losing too much information.
\end{remark}

\begin{remark}
Assuming the true model is $y=\sum_{i=1}^p\beta_i x_i$, then, after doing AS it becomes $y=\sum_{i=1}^p \tilde{x}_i$. In this case, AS just makes  all dimensions of features into the measure of $y$, in other word, the change of $\tilde{x}_i$ will reflect onto $y$ directly.
\end{remark}

\begin{remark}
The coefficients of OLS have gained the information about variance of each feature as well as their covariance, so AS eliminates the effect of variance and reweight features according to their covariance and their contributions to $y$. Therefore, in linear case, AS does Standardization and then adds some prior weights on raw data.
\end{remark}

\subsection{Generalized Adaptive Scaling (GAS)}
Assume we have raw data $(X,Y)$ and we separate it into training data $({X}_{training},Y_{training})$ and testing data$({X}_{testing},Y_{testing})$. And then we do Generalized Adaptive Scaling on both training data and testing data.

Algorithm:\\
 1. Center Data.
 $$X=X-\bar{{X}}_{training}$$\\
 2. Apply OLS on training data to get coefficients.
 $$\beta = (X_{training}^T X_{training})^{-1}X_{training}^T Y_{training}$$\\
 3. Multiple the coefficients to the raw data.
 $${X'}_i=|\beta_i|^\gamma \cdot X_i$$\\
 4. Use ${X'}_{training}$ and $Y_{training}$ to learn the model and take $\gamma$ as a hyper-parameter.\\
 5. Use ${X'}_{testing}$ and $Y_{testing} $ to validate the model.
 
\begin{remark}
Here $\gamma$ is a parameter between 0 and 1, and which can be learned via cross-validation. When $\gamma = 0$ GAS does not scale features, and when $\gamma = 1$ GAS becomes AS.
\end{remark}

\begin{remark}
$\gamma$ adjusts the weight of variance and prior, the bigger the $\gamma$ is, the more prior will be applied onto raw data.
\end{remark}
 
\subsection{Adaptive Scaling for High Dimensional Data (ASH)}
As for High Dimensional Data with p$\gg$n, OLS does not work, but we can apply OLS on each dimension of data. So we can have Adaptive Scaling for high dimension data:

Algorithm:\\
 1. Center Data.
 $$X=X-\bar{{X}}_{training}$$\\
 2. Apply OLS on each dimension of training data separately to get the coefficient.
 $$\beta_i = (X_{i,training}^T X_{i,training})^{-1}X_{i,training}^T Y_{training}, i=1...p$$\\
 3. Multiple the coefficients to the raw data.
 $${X'}_i=|\beta_i|^\gamma \cdot X_i$$\\
 4. Use ${X'}_{training}$ and $Y_{training}$ to learn the model and take $\gamma$ as a parameter.\\
 5. Use ${X'}_{testing}$ and $Y_{testing} $ to validate the model.
 
\begin{remark}
Since we do OLS separately, AS for high dimension data does not consider the covariance of features which is the main difference from the previous two methods. However, that also makes ASH more robust than the former two.
\end{remark}

\begin{remark}
Still, one can use cross-validation to get the best value of $\gamma$. One can also just set $\gamma=1$ which has more explanation for the coefficient.
\end{remark}

\begin{remark}
This idea is alike the thought of ensemble.........
\end{remark}

\begin{remark}
Since we deal with every dimension of features separately, so we can apply any other models, such as spline, to fit one-dimension predictors and then combine them together. However, by using more complicated models, more parameters will be introduced at the same times, which require more data to learn them. Indeed, as for classification, it is just the FANS method if we set $\gamma=1$, and take Naive Bayes Classifor or log likelihood ratio instead of linear regression. 
\end{remark}

\section{Scaling Effects on Models}\label{affection}
In this section, we will focus on different statistical learning methods' sensitivities to feature scaling. Some models such as, Naive Bayes Classifier, Tree models and linear regression, will not be influenced by different scaling methods. However, most models are sensitive to data preprocessing methods. For distance based methods, like K-NN, K-Means and so on, proper scaling methods prevent features with initial large range or variance dominate other features. Scaling also impacts some optimizing algorithm like gradient decent and so on. Ordinary least squares and LASSO are given as two examples.

Here is the basic setting for this section: 
Let $X$ be the raw data with $n$ observations and $p$ dimensions, and $\hat{X}$ be the scaled data with $\hat{X_i} = \alpha_i (X_i-\mu_i), i=1\cdots p$ where $\alpha_i$  and $\mu_i$ are all real number.

\subsection{Scale-invariant Models}
Although feature scaling is a big issue in most cases, there are some methods which will not be impacted by scaling, so we call them scale-invariant models. Here are some examples:

1. Likelihood Based Models:
Most likelihood based models which use likelihood ratio or log of likelihood ratio as new feature to do analysis make all features in the same probability measurement, so these models will be scale-invariant.

Naive Bayes Classifier:
Naive Bayes Classifier is a typical instance of this kind of models. 
$$\hat{y}=\mathop{\arg\max}_{k}p(C_k)\prod_{i=1}^p p(x_i|C_k).$$
Since $p(C_k)$ is prior which only determined by $y$ value, and $p(x_i|C_k)$ is estimated by likelihood estimation which will not be affected by different scaling methods.
Overall, Naive Bayes Classifier is scale-invariant model.

Feature Augmentation via Nonparametircs and Selection(FANS):
Feature Augmentation via Nonparametircs and Selection(FANS)\cite{Jianqing Fan} uses marginal density ratio estimates to transform the original data and makes all features in the unit of likelihood ratio. So, similar to Naive Bayes Classifier, it is scale-invariant model. Actually, FANS translates the original features to $\log \frac{f_i(x_i)}{g_i(x_i)}$, and then, applies statistical learning models like logistic regression with or without penalty,  SVM and so on using the translated data.

2. Tree-based Models:
Tree-based models like CART Decision Tree, Random Forrest and so on, use impurity measure $Q_m(T)$ for model building\cite{learning}. Whether using Misclassification erro, Gini index or Cross-entropy, the impurity measure is determined by $\hat{y}$ and $y$ here $\hat{y}$ is the estimated label, and will not affected by the scale of $X$. So generally speaking, tree-based models are also scale-invariant models.

3. Linear Regression:
Ordinary Least Squares Linear Regression(OLS) is the most basic statistical model and it is scale-invariant for prediction. 

OLS estimator is
 $$\hat{\beta}^{ols} = \mathop{\arg\min}_{\beta} \ \ \| Y-X\beta\|^2 = (X^TX)^{-1}X^TY.$$
 Its predicted value is
  $$\hat{Y}=X\hat{\beta}^{ols} = \hat{\beta}_1^{ols} x_1 + \hat{\beta}_2^{ols} x_2 + ... + \hat{\beta}_p^{ols}x_p.$$
 After scaling OLS estimator is 
  $$\tilde{x}_i=x_i/c_i, \tilde{X}=(\tilde{x}_1,\tilde{x}_2,...,\tilde{x}_p).$$
  $$\hat{\tilde{\beta}}^{ols} = \mathop{\arg\min}_{\beta} \ \ \| Y-\tilde{X}\beta\|^2 = (\tilde{X}^T\tilde{X})^{-1}\tilde{X}^TY.$$
  $$\hat{\tilde{\beta}}_i^{ols}=c_i\hat{\beta}_i.$$
 Its predicted value is
  $$\hat{\tilde{Y}}=\tilde{X}\hat{\tilde{\beta}}^{ols} = \hat{\tilde{\beta}}^{ols} \tilde{x}_1 + \hat{\tilde{\beta}}^{ols} \tilde{x}_2 + ... + \hat{\tilde{\beta}}^{ols}\tilde{x}_p=\hat{Y}.$$
So OLS is scale-invariant model if we only consider its prediction.

\subsection{Scale-variant Models}
1. Distance Measure Based Methods:
Algorithms using distance measures such as K-NN, K-Means will be affected seriously by feature scaling. Take Euclidean distance measure for example:

Square of distance between two observation $X_i$ and $X_j$
$$d_{ij}^2 = \sum_{k=1}^p(x_{i,k}-x_{j,k})^2.$$
After scaling, square of distance between two observation $X_i$ and $X_j$ become
$$\tilde{x}_i=x_i/c_i, \tilde{X}=(\tilde{x}_1,\tilde{x}_2,...,\tilde{x}_p).$$
$$\tilde{d}_{ij}^2=\sum_{k=1}^p(\tilde{x}_{i,k}-\tilde{x}_{j,k})^2= \sum_{k=1}^p(x_{i,k}-x_{j,k})^2\frac{1}{c_k^2}.$$ 
Usually $d_{ij}$ is not equal to $\tilde{d}_{ij}$, and if there exist some features with large norm then $d_{ij}$ will be dominated by them.
Therefor, distance measure based methods always are scale-variant.
 
2. Models with Feature Selection or Shrinkage:
In order to handle high-dimension issue, many models with feature selection or coefficient shrinkage such as LASSO\cite{LASSO}, Adaptive LASSO\cite{Adaptive LASSO}, SCAD\cite{SCAD}, MCP\cite{MCP}, Ridge Regression\cite{Ridge Regression} are proposed. Most of them are sensitive to scale of features. Take LASSO as an example.

LASSO minimizes:

 \begin{eqnarray*}
g(\beta)&=&\frac{1}{2} \| Y-X\beta\|^2+\lambda \|\beta\|_1\\
            &=&\frac{1}{2} \| Y-\tilde{X}\tilde{\beta}\|^2+\lambda \sum_{i=1}^p |\beta_i|\\
            &=&\frac{1}{2} \| Y-\tilde{X}\tilde{\beta}\|^2+\lambda \sum_{i=1}^p |\tilde{\beta}_i|/c_i\\
            &=&\frac{1}{2} \| Y-\tilde{X}\tilde{\beta}\|^2+\sum_{i=1}^p|\tilde{\beta}_i|\omega_i\\
            &=&C+\sum_{i=1}^p(\frac{1}{2}(\tilde{\beta}_i-\hat{\tilde{\beta}}_i^{ols})^2+\omega_i |\tilde{\beta}_i|),
 \end{eqnarray*}
 where 
 $\hat{\tilde{\beta}}^{ols}=(\tilde{X}^T\tilde{X})^{-1}\tilde{X}^TY, \omega_i=\frac{\lambda}{c_i}$ and if we aume $ x_i\perp x_j, i\neq j$ then

 \begin{eqnarray*}
\hat{\tilde{\beta}}^{lasso} &=& \mathop{\arg\min}_{\beta} \ \ g(\beta) \\
                                        &=& \mathop{\arg\min}_{\beta} \ \ \sum_{i=1}^p(\frac{1}{2}(\tilde{\beta}_i-\hat{\tilde{\beta}}_i^{ols})^2+\omega_i |\tilde{\beta}_i|). \\
 \end{eqnarray*}
The KKT conditions are
 \begin{eqnarray*}
(\tilde{\beta}_i-\hat{\tilde{\beta}}_i^{ols})+\omega_i sgn(\tilde{\beta}_i)&=&0 \ \ \ if \tilde{\beta}_i\neq 0\\
|\hat{\tilde{\beta}}_i^{ols}|&\le & \omega_i \ \ \ if \tilde{\beta}_i \neq 0.
 \end{eqnarray*}
Coefficients of LASSO:
  \begin{eqnarray*}
\hat{\tilde{\beta}}_j^{lasso}& = &sgn(\hat{\tilde{\beta}}_j^{ols})(|\hat{\tilde{\beta}}_j^{ols}|-\frac{\lambda}{c_j})_+\\
& = &sgn(\hat{\tilde{\beta}}_j^{ols})(\frac{\frac{c_j^2}{\sigma_i^2} |x_j^TY|-\lambda}{c_j})_+.
  \end{eqnarray*}
If $\|x_j\|$ is small, then possibly $\hat{\tilde{\beta}}^{lasso}_j=0.$\\
If truth is $\beta_j \neq 0$, then we missed a true feature.\\
If $\|x_j\|$ is large, then possibly $\hat{\tilde{\beta}}_j^{lasso}\neq 0.$\\
If truth is $\beta_j = 0$, then we wrongly selected a fake feature.

So far, we have showed that LASSO is scale-variant. Although most feature selection or coefficient shrinkage methods are scale-variant, they have different levels of sensitivity to feature scaling which will be shown in simulation in section \ref{simulation}.

3. Gradient Descent:
Many statistical learning methods such as logistic regression\cite{learning}, Linear Discriminant Analysis(LDA)\cite{learning}, Support Vector Machine(SVM)\cite{learning},Neural Networks(NN)\cite{Prasad} use Gradient Descent to find the optimal parameters. Feature scaling will affect the performance of Gradient Descent algorithm so it will impact the performance of these models.

Indeed, for gradient descent algorithm:\\
$$b_{j+1}:=b_j+\delta b_j$$
$$\delta b_j = \gamma \sum_{i=1}^n(\hat{y}^{(i)}-y^{(i)})x_j^i,$$
here $\gamma$ is the learning rate which is the same for all dimensions of data, $\hat{y}$ is the estimated value and $y$ is the original one, so features with large scaling will have bigger $\delta b_j$. In other word, all features are in different learning rate if not scaled properly.

So far, we have seen some examples of scale-invariant models as well as scale-variant models. Although some models are scale-invariant, it will not hurt if we scale the data before learning models. As for scale-variant models, they have different levels of sensitivity to feature scaling, and this can also be seen in both simulation and empirical study. Since proper scaling is necessary for many models, which method to do feature scaling becomes a great issue now.

\section{Simulation}\label{simulation}
In this section, we apply the feature scaling methods mentioned in section \ref{intro} as well as  Adaptive Scaling and Generalized Adaptive Scaling on a linear setting simulation to analyses their performance on different regression methods.

\subsection{Simulation1}
We set the true model as $Y=\beta X + \epsilon.$ We take $n=100, p=8$ and $\beta=(3,1.5,0,0,2,0,0,0)$ which means having some silence features. The predictors $x_i (i = 1,...,n)$ are iid normal vectors. We set the pairwise correlation between $x_{j_1}$ and $x_{j_2}$ to be $cor(j_1,j_2) = {0.5}^{|j_1-j_2|}$. Indeed, the setting is the same as one of simulations in \cite{Adaptive LASSO}. And we take $\epsilon$ from normal distribution $N(0,3^2)$.

For each dataset ($n=100$), we use half data as training data($n_{training}=50$) and the left as testing data. We use all the  scaling methods mentioned before both on training and testing data. We don't use Gelman's method because there is no binary data in this case. After Scaling, we apply LASSO, Adaptive LASSO, SCAD, MCP and Garrote on the training data to learn models. Five-fold cross validation is applied when needed. We repeat 100 times for each result.

We report the relative prediction error (RPE),$RPE=E[(Y-\beta X)^2]/\epsilon^2$.

Mean RPE (the smaller the better):\\
\begin{table}[H]
\begin{center}
\begin{tabular}{ | l | l | l | l | l | l | }
\hline
	& LASSO & Adaptive LASSO & Garrote & SCAD & MCP  \  \\ \hline
	No & 1.2511 & 1.119 & 1.1589 & \textbf{1.1393} & \textbf{1.1328} \\ \hline

	AS& \textbf{1.2395} & 1.119 & 1.1589 & 1.1599 & 1.1633 \\ \hline
	GAS& 1.2398 & 1.119 & 1.1589 & 1.1656 & 1.1657 \\ \hline
		ASHD & 1.2516 & 1.119 & 1.1589 & 1.1528 & 1.1504 \\ \hline
		Stand & 1.2439 & 1.119 & 1.1589 & 1.1736& 1.1686 \\ \hline	
	RS & 1.2457 & 1.119 & 1.1589 & / & / \\ \hline
	PS& 1.2464 & 1.119 & 1.1589 & 1.1762 & 1.1736 \\ \hline
	VS& 1.5182 & 1.2431 & 1.2911 & /& / \\ \hline
	LS& 1.3361& 1.119 & 1.1589 & 1.2000 & 1.1945 \\ \hline
\end{tabular}
\end{center}
\caption{No=No Scaling, AS=Adaptive Scaling, GAS=Generalized Adaptive Scaling,  ASHD=Adaptive Scaling for High-dimensional, Stand=Standardization, RS=Range Scaling, PS=Pareto Scaling, VS=Vast Scaling, LS=Level Scaling.}
\end{table}

\begin{remark}
The result shows that, no single feature scaling method always performs best on all regression models. Overall, Adaptive Scaling performs well among scaling methods and is better than Standardization and Range Scaling.
\end{remark}

\begin{remark}
This result also shows that different statistical learning models have different sensitivity to feature scaling. In this case, the order of sensitivities of these 5 regression methods is $LASSO>MCP,SCAD>Garrote,Aaptive Lasso$.
\end{remark}

\subsection{Simulation2}
In this simulation, we mainly focused on the scaling impact on LASSO. Still, we set the true model as $Y=\beta X + \epsilon.$ We take $n=1000, p=8$ but let $\beta$ take different pattern and we focus on the variable selection performance by different scaling methods followed by LASSO. The training ration is 0.5 and each simulation is repeated 100 times to give out the result. We take $\epsilon$ from normal distribution $N(0,5^2).$

We report the fake selection ration, which means the ration of wrongly picking the silence variables out; lost selection ration, which means the ration of lost the true variables and RPE which is introduced in Simulation1.

Firstly, we consider the independent cases.  $X$ comes from $N(\mu, \Sigma)$, here $\mu=0$, $\Sigma_{i,j}=0,\ for \  i \neq j$ and $\Sigma_{i,i}=3^i.$

\begin{case} 
$\beta=(0,0,0,0,1,1,1,1).$
\begin{table}[H]
\begin{tabular}{ | l | l | l | l | l | l | l | l | l |}
\hline
   & No & AS & ASHD & Stand & PS &RS &VS &LS  \ \\ \hline
Fake rat &3.75\%  & 0 &26.00\%  &30.50 \% & 10.00\%& 29.75\% &70.25\% &71.00\%\\ \hline
Lost rat &0  &0  &1.00\%  &0  &0 &0  &0.50\% &0.50\%\\ \hline
PRM &4.956  &\textbf{4.949}  &5.414  &5.082  &4.987 &5.069  &13.012 &13.485\\ \hline
\end{tabular}
\end{table}

\end{case}

\begin{case}
$\beta=(1,1,1,1,0,0,0,0).$
\begin{table}[H]
\begin{tabular}{ | l | l | l | l | l | l | l | l | l |}
\hline
   & No & AS & ASHD & Stand & PS &RS &VS &LS  \ \\ \hline
Fake rat &98.75\%  &22.25\% &36.75\%  &52.50 \% & 84.75\%& 52.25\% &63.00\% &65.25\%\\ \hline
Lost rat &3.00\%  &5.25\%  &5.25\% &1.25\%  &4.00\%  &1.00\% &4.50\%  &3.75\%\\ \hline
PRM &1.249  &\textbf{1.208}  &  &1.214 &1.213 &1.239  &1.354 &1.359\\ \hline
\end{tabular}
\end{table}
\end{case}

\begin{case}
$\beta=(0,0,1,1,1,1,0,0).$
\begin{table}[H]
\begin{tabular}{ | l | l | l | l | l | l | l | l | l |}
\hline
   & No & AS & ASHD & Stand & PS &RS &VS &LS  \ \\ \hline
Fake rat &59.00\%  &14.25\% &37.25\%  &56.00 \% & 58.50\%& 57.00\% &72.25\% &71.50\%\\ \hline
Lost rat &0  &0 &0.25\% &0 &0  &0 &0.75\%  &1.75\%\\ \hline
PRM &1.673  &\textbf{1.647}  &1.669 &1.682 &1.675  &1.683 &2.811 &2.898\\ \hline
\end{tabular}
\end{table}
\end{case}

\begin{case}\label{bad_case1}
$\beta=(1,1,0,0,0,0,1,1).$
\begin{table}[H]
\begin{tabular}{ | l | l | l | l | l | l | l | l | l |}
\hline
   & No & AS & ASHD & Stand & PS &RS &VS &LS  \ \\ \hline
Fake rat &4.50\%  &0 &11.25\%  &24.25 \% & 18.25\%& 24.50\% &65.25\% &63.75\%\\ \hline
Lost rat &48.75\%  &49.00\%&32.50\% &9.50\% &34/75\%  &10.00\% &13.50\%  &13.50\%\\ \hline
PRM &5.180  &5.152  &5.016 &\textbf{4.976} &5.076 &4.976 &14.361 &16.283\\ \hline
\end{tabular}
\end{table}
\end{case}

Then, we consider the correlate cases.  $X$ comes from $N(\mu, \Sigma)$, here $\mu=0$, $\Sigma_{i,j}=0.5^{|i-j|},\ for \  i \neq j$ and $\Sigma_{i,i}=3^i.$

\begin{case}
$\beta=(0,0,0,0,1,1,1,1).$
\begin{table}[H]
\begin{tabular}{ | l | l | l | l | l | l | l | l | l |}
\hline
   & No & AS & ASHD & Stand & PS &RS &VS &LS  \ \\ \hline
Fake rat &3.75\%  &0 &25.25\%  &30.00 \% & 10.00\%& 29.75\% &70.00\% &71.25\%\\ \hline
Lost rat &0  &0 &1.00\% &0 &0  &0 &0.50\%  &0.50\%\\ \hline
PRM &4.957  &\textbf{4.950}  &5.410 &5.080 &4.988  &5.073 &13.026 &13.458\\ \hline
\end{tabular}
\end{table}
\end{case}

\begin{case}
$\beta=(1,1,1,1,0,0,0,0).$
\begin{table}[H]
\begin{tabular}{ | l | l | l | l | l | l | l | l | l |}
\hline
   & No & AS & ASHD & Stand & PS &RS &VS &LS  \ \\ \hline
Fake rat &98.00\%  &21.50\% &30.50\%  &51.50\% & 85.75\%& 52.25\% &65.25\% &66.00\%\\ \hline
Lost rat &3.50\%  &5.75\% &4.50\% &1.25\% &3.50\%  &0.75\% &4.25\%  &3.50\%\\ \hline
PRM &1.273  &\textbf{1.220}  &1.225 &1.231 &1.263  &1.375 &1.375 &1.382\\ \hline
\end{tabular}
\end{table}
\end{case}

\begin{case}
$\beta=(0,0,1,1,1,1,0,0).$
\begin{table}[H]
\begin{tabular}{ | l | l | l | l | l | l | l | l | l |}
\hline
   & No & AS & ASHD & Stand & PS &RS &VS &LS  \ \\ \hline
Fake rat &60.25\%  &14.00\% &37.00\%  &54.00 \% & 57.50\%& 53.25\% &72.50\% &72.75\%\\ \hline
Lost rat &0  &0 &0.25\% &0 &0  &0 &0.5\%  &1.50\%\\ \hline
PRM &1.686 &\textbf{1.661}  &1.713 &1.697 &1.687  &1.697 &2.781 &2.859\\ \hline
\end{tabular}
\end{table}
\end{case}

\begin{case}\label{bad_case2}
$\beta=(1,1,0,0,0,0,1,1).$
\begin{table}[H]
\begin{tabular}{ | l | l | l | l | l | l | l | l | l |}
\hline
   & No & AS & ASHD & Stand & PS &RS &VS &LS  \ \\ \hline
Fake rat &5.25\%  &0 &11.25\%  &24.00 \% & 22.75\%& 24.75\% &64.75\% &62.00\%\\ \hline
Lost rat &48.75\% &49.00\% &30.50\% &9.25\% &31.75\% &9.25\% &13.00\% &14.25\%\\ \hline
PRM &5.217 &5.191  &5.026&4.971 &5.059  &\textbf{4.963} &15.192 &16.378\\ \hline
\end{tabular}
\end{table}
\end{case}

\begin{conclusion}
Adaptive Scaling improve both the feature selection performance and PRM of LASSO quite a lot in most cases, expect when we have features with bot extremely small and large deviation together( Case \ref{bad_case1} and Case \ref{bad_case2}).
\end{conclusion}
\section{Empirical Study}\label{empirical}
In this section, we will look at one real data set for classification to compare the performance of all these scaling methods. We use commonly used classification models like Logistic Regression, SVM, Neural Network and so on to do prediction after feature scaling.

\subsection{Data}
We use a data set from UCI named Default of credit card clients Data Set\cite{credit data}. This payment data recorded the default of credit card clients and wanted to predict future default by using historical recording. The data set has 30,000 instances and 24 attributes, in other word, in this case, n=30,000, p=24. Here are two examples of instances:

\begin{table}[ht]
\tiny
\begin{tabular}{ | l | l | l | l | l | l | l | }
\hline
	ID & LIMIT\_BAL & SEX & EDUCATION & MARRIAGE & AGE & PAY\_0 \\ \hline
	1 & 20000 & 2 & 2 & 1 & 24 & 2 \\ \hline
	2 & 120000 & 2 & 2 & 2 & 26 & -1 \\ \hline
	 &  &  &  &  &  &  \\ \hline
	ID&PAY\_6 & BILL\_AMT1 & BILL\_AMT6 & PAY\_AMT1 & PAY\_AMT6 & default payment \  \\ \hline
	1&-2 & 3913 & 0 & 0 & 0 & 1  \  \\ \hline
	2&2 & 2682 & 3261 & 0 & 2000 & 1 \  \\ \hline
\end{tabular}
\caption{Two instances}
\end{table}

In this case we have both numerical data like age, loan limit and payment, and we also have categorical data (sex, marriage) as well as ordinal data (education). So it is more important for us to do feature scaling properly. We use half data as training data ($n_{training}=15,000$) and the left as testing data. We use all the  scaling methods mentioned before both on training and testing data. After Scaling, we apply K-NN, LDA, K-Means, Tree C5.0, Naive Bayes, Logistic Regression, Neural Network and SVM on the training data to learn models. Five-fold corrs-validation is applied when needed. We repeat 100 times on K-NN and K-means because both of them are very sensitive to feature scaling. We repeat 10 times on the rest models considering their less sensitivities and they are time consuming.

\subsection{Result and Analysis}
The table below shows average accuracy of different models with different feature scaling methods.\\
 \begin{table}[ht]
 \tiny
\begin{tabular}{ | l | l | l | l | l | l | l | l | l | }
\hline
	       &K-NN & LDA & K-Means & Naive Bayes & C5.0 & Logistic Reg & Neural Net & SVM  \\ \hline
	No & 0.7641 & 0.8111 & 0.6918 & 0.7936 & 0.8197 & 0.8105 & 0.7785 & 0.8071\\ \hline
	Stand& 0.8030 & 0.8111 & 0.6814 & 0.7936 & 0.8197 & 0.8105 & 0.8187 & 0.8069\\ \hline
	RS & 0.8027 & 0.8111 & 0.5388 & 0.7936 & 0.8197 & 0.8105 & 0.8151 & 0.8069\\ \hline
	AS & 0.8087 &0.8111 & 0.5833 & 0.7936 & 0.8197 & 0.8105 & 0.8201 & 0.8072\\ \hline
    GAS& 0.7714 & 0.8111 & 0.6949 & 0.7936 & 0.8197 & 0.8105 & 0.8171 & 0.8068\\ \hline
    ASHD& 0.8079 & 0.8111 & 0.5493 & 0.7936 & 0.8197 &\textbf{0.8111} &\textbf{0.8215} & \textbf{0.8090}\\ \hline
	PS& 0.7708 & 0.8111 & \textbf{0.6953} & 0.7936 & 0.8197 & 0.8105 & 0.7852 & 0.8071\\ \hline
	VS& 0.7779 & 0.8111 & 0.5723 &  0.7936& 0.8197& 0.8105 & 0.8164 & 0.8072\\ \hline
	LS & \textbf{0.8092\textsc{•}} & 0.8111 &0.6730 & 0.7936 & 0.8197& 0.8105 & 0.8103 & 0.8071\\ \hline
\end{tabular}
\caption{Accuracy of Prediction}
\end{table}

Here is the Standard Deviation of the Accuracy above.
 \begin{table}[ht]
 \tiny
\begin{tabular}{ | l | l | l | l | l | l | l | l | l | }
\hline
	       &K-NN & LDA & K-Means & Naive Bayes & C5.0 & Logistic Reg & Neural Net & SVM  \\ \hline
	No& 0.0025 & 0.0022 & 0.0031 & 0.0041 & 0.0021 & 0.0026 & 0.0026 & 0.0048\\ \hline
	Stand& 0.0020 & 0.0022 & 0.0051 & 0.0041 & 0.0021 & 0.0026 & 0.0024 & 0.0052\\ \hline
	RS & 0.0021 & 0.0022 & 0.0241 & 0.0041 & 0.0021 & 0.0026 & 0.0132 & 0.0049\\ \hline
	AS & 0.0028 & 0.0022 & 0.0754 & 0.0041 & 0.0021 & 0.0026 & 0.0020 & 0.0049\\ \hline
    GAS& 0.0020 & 0.0022 & 0.0029 & 0.0041 & 0.0021 & 0.0026 & 0.0023 & 0.0054\\ \hline
    ASHD& 0.0025 & 0.0022 & 0.1249 & 0.0041 & 0.0021 & 0.0026 & 0.0030 & 0.0052\\ \hline
	PS & 0.0017 & 0.0022 & 0.0028 & 0.0041 & 0.0021 & 0.0026 & 0.0144 & 0.0048\\ \hline
	VS& 0.0030 & 0.0022 & 0.0060 & 0.0041 & 0.0021 & 0.0026 & 0.0038 & 0.0048\\ \hline
	LS & 0.0026 & 0.0028 & 0.1019 & 0.0041 & 0.0021 & 0.0528 & 0.0026 & 0.0038\\ \hline
\end{tabular}
\caption{Standard Deviation of Accuracy}
\end{table}

From the result, we can have following conclusion.
\begin{conclusion}
Just like the result of simulation, different feature scaling methods have different influence in most statistical learning models and none  always performs better than others. However, new scaling methods, Adaptive Scaling, Generalized Adaptive Scaling and Adaptive Scaling for High Dimensional Data, have great overall performance, and again, they are much better than Standardization and Rang Scaling, the two widely used methods.
\end{conclusion}

\begin{conclusion}
In this example, Tree model does a good job and is scale-invariant. However, Adaptive Scaling for High Dimensional Data followed by Neural Network has the highest accuracy (82.15\%). 
\end{conclusion}

\begin{conclusion}
Although some models do not perform well without scaling or with improper scaling, like K-NN, K-Means and Neural Network, if we can apply proper scaling methods before them, they can also approach the best performance.
\end{conclusion}

\begin{conclusion}
Finally, in this case, the order of sensitivities of models is K-NN, K-Means $>$ Neural Network, SVM $>$ Logistic Regression $>$LDA, Naive Bayes, Tree C5.0.
\end{conclusion}

\section{Conclusion and Discussion}\label{conclusion}
In this paper, we mainly focus on two fundamental objects. Firstly, we point out that it is necessary to apply proper feature scaling on raw data for most statistical learning models including K-NN, K-means, Neural Network, SVM and so on. Both simulation and empirical study also show that proper scaling will raise performances of most models.

Secondly, we propose three new scaling methods. We use ordinary least squares regression's coefficients to eliminate the variance of features and add prior weights on them at the same time. We also generalize Adaptive Scaling and fit it to high dimensional data. Advantages of our methods are shown both in simulation and in real data analysis.

Although we see good performance of Adaptive Scaling method, more work need be done in the future. First of all, more theoretical results should be built about this new method. What is more, more real data sets can be tested in the future. At last, more statistical learning models can be tried to verify the robustness of our methods.

\
\appendix

\section{Related Works}
Here is a list of commonly used scaling methods with some comments:

1. Standardization (Z-Score transformation, Autoscaling, Normalization):

$$x'=\frac{x-\bar{x}}{\sigma}$$

where $\bar{x}$ is the mean of $x$ and $\sigma$ is its standard deviation.

Motivation: Make all the features' variance to be 1.
 
Advantages: All features become roughly equally important. The measure problem is eliminated. Unlike Range Scaling, Standardization is not so sensitive to outliers.
 
Disadvantages: Features will not be in the same range. The location and scale information of the original data is lost\cite{Jain A K}. As for categorical data with many levels, the variance will be quite small if we introduce dummy variables, so we will divide a tiny number on these features which will make them less likely to be selected in some statistical learning methods such as LASSO.
 
2. Range Scaling(Min-Max Scaling):

 $$x'=\frac{x-min(x)}{max(x)-min(x)}.$$
 
Motivation: Make all the features in the same range $[0,1].$
 
Advantages: All the features are made roughly equally important. Measuring problem is eliminated.
 
Disadvantages: Very sensitive to outliers. If the new data fall outside the old ones, the model has to be renewed to fix this issue.

3. Pareto Scaling:

$$x'=\frac{x-\bar{x}}{\sqrt{\sigma}}.$$

Motivation: Reduce the impact of variance. 
 
Advantages: Compared to standardization, Pareto Scaling is closer to the original measure.
 
Disadvantages: Features are neither in the same range, nor have unit variance. Although it is better than Standardization on the problem of categorical data with many levels, it still has some bias in such issue.
 
4. Scaling regression inputs by dividing by two standard deviations\cite{Gelman A}:

$$\begin{cases}
x'=\frac{x-\bar{x}}{2\sigma•} \text{ if } x \text{ is a numerical feature}\\ 
x'=x \text{ if } x \text{ is a dummy variable from a categorical feature}  
\end{cases}.$$

Motivation: Design for the interpretation of regression coefficients. 

Advantages:  Using binary data as benchmark makes the meaning of regression coefficients comparable both for numerical features and categorical features.
 
Disadvantages: Ignore the variance of binary data. This method sometimes solves the issue of categorical data with many levels, but in the same time lightening  the importance of numerical features also introduce bias. What is more, it is proposed mainly for linear regression, so the effect on other models is unclear.

5. Vast Scaling:

$$x'=\frac{x-\bar{x}}{\sigma}\cdot \frac{\bar{x}}{\sigma}.$$

Motivation: Enlarge the features with small fluctuations.
 
Advantages: By enlarging the features with small fluctuations, one can see their impact more clearly in some statistical learning models.
 
Disadvantages: Features are neither in the same range, nor have unit variance. Not suitable if there is no prior knowledge to focus on the features with small variance. Moreover, centering is applied before scaling, then all features will be roughly equal to 0, and this loses most of information of the raw data, in another word, this method is sensitive to both variance and mean.
 
6. Level Scaling:

$$x'=\frac{x-\bar{x}}{\bar{x}}.$$

Motivation: Use mean to measure different features.
 
Advantages:  Insensitive to features' variance. Features will not be in the same range. 
 
Disadvantages: Features are neither in the same range, nor have unit variance. If the mean is close 0, then this method will be unstable.
 
\end{document}